\pdfoutput=1

\documentclass[11pt]{article}

\usepackage[preprint]{acl}
\usepackage{booktabs}
\usepackage[many]{tcolorbox}

\usepackage{times}
\usepackage{latexsym}
\usepackage{comment}

\usepackage[T1]{fontenc}

\usepackage[utf8]{inputenc}

\usepackage{microtype}

\usepackage{inconsolata}

\usepackage{graphicx}
\usepackage{authblk}

\title{Pause-Tuning for Long-Context Comprehension: A Lightweight Approach to LLM Attention Recalibration} 

\author{
    \textbf{James Begin} \quad
    \textbf{Namit Agrawal} \quad
    \textbf{Eshan Singh} \\
    \textbf{Yicheng Fu} \quad
    \textbf{Sean O’Brien} \quad
    \textbf{Vasu Sharma} \quad
    \textbf{Kevin Zhu}
}
\affil{Algoverse AI Research}
\affil{\texttt{kevin@algoverse.us, sean@algoverse.us}}

\date{}

\begin{document}
\maketitle
\begin{abstract}
LLMs have demonstrated remarkable proficiency in understanding tasks but continue to struggle with long-context comprehension, particularly with content located in the middle of extensive inputs. This limitation, known as the Lost-in-the-Middle (LITM) problem, hinders models from fully processing and utilizing information across lengthy contexts. To address this issue, we introduce pause-tuning, a technique that redistributes attention to enhance comprehension of long-context inputs. Our approach involves fine-tuning language models on datasets with artificially inserted pause tokens, which serve to segment the input into smaller, more manageable parts. We evaluate pause-tuning against alternative approaches using the Needle-in-a-Haystack benchmark, where models must retrieve information embedded within contexts of up to 128K tokens. Experimental results demonstrate significant performance gains, with the LLaMA 3.2 3B Instruct model and the LLaMA 3.1 8B Instruct model improving by 10.61\% and 3.57\% respectively on average, suggesting that pause-tuning successfully enhances attention redistribution and improves long-context retention. The code and data are
available at \url{https://anonymous.4open.science/r/LITM-PauseTokens-7357}.

\end{abstract}

\section{Introduction}

Language models like GPT~\cite{brown2020languagemodelsfewshotlearners} and LLaMA~\cite{grattafiori2024llama3herdmodels} have demonstrated remarkable utility in tasks such as summarization, long document analysis, and contextual understanding \citep{minaee2024largelanguagemodelssurvey}. Effectively handling long contexts is essential for maintaining the accuracy and reliability of a model’s output in these applications. However, language models often suffer from the lost-in-the-middle problem \citep{liu-etal-2024-lost}, where they disproportionately focus on the beginning and end of sequences while neglecting information in the middle. Existing attempts at solutions often fall short of being broadly applicable. Many approaches rely on computationally intensive mechanisms or involve modifications of the base language model other than simple fine-tuning \citep{he2024lostmiddlemasteringlongcontext, tworkowski2023focusedtransformercontrastivetraining, liu2023blockwiseparalleltransformerlarge}. While effective in certain scenarios, these methods may be impractical in resource-constrained environments or general-purpose applications.

\begin{figure}
    \centering
    \includegraphics[width=1\linewidth]{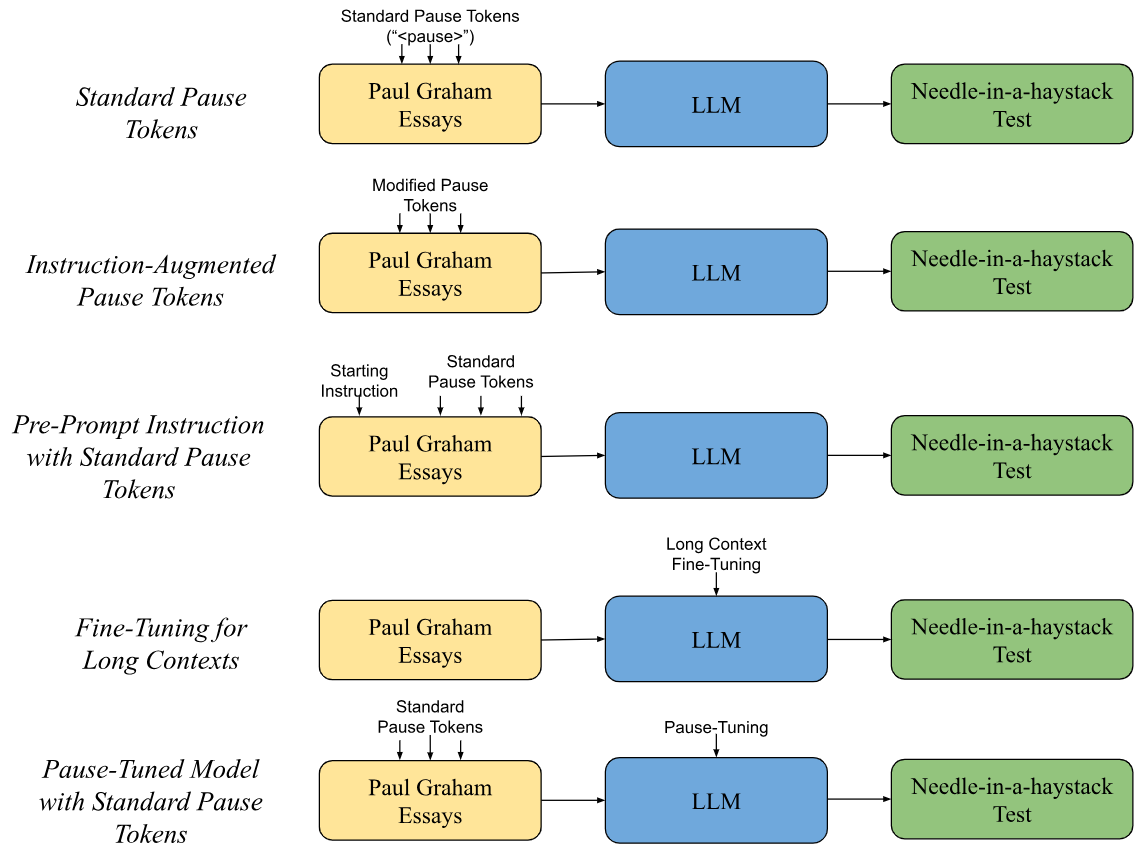}
    \caption{We propose five potential techniques for pause token injection and test them on the needle-in-a-haystack evaluation framework.}
    \label{fig:pause_tuning}
\end{figure}

To address this gap, we propose a novel approach that utilizes pause tokens \citep{goyal2024thinkspeaktraininglanguage} to mitigate the LITM problem. Pause tokens are markers that are strategically inserted into the input sequence, intended to recalibrate the model's attention distribution. These tokens prompt the model to pause and process information before proceeding with the rest of the sequence. By segmenting the input sequence into smaller, more manageable chunks, pause tokens allow the model to process each segment with greater focus and parity. This simple yet effective method offers a lightweight alternative to existing resource-intensive techniques. We investigate various strategies for inserting pause tokens, as depicted in ~\autoref{fig:pause_tuning}.

We evaluate the efficacy of pause tokens through a series of needle-in-a-haystack experiments \citep{needleinahaystack2023, nelson2024needlehaystackmemorybased}, comparing base models against those fine-tuned for pause-aware long-context processing. Our results demonstrate that:
\begin{itemize} 
    \item Pause-tuning consistently improves long-context retention and processing, outperforming alternative techniques.
    \item Pause tokens induce meaningful shifts in attention distribution, enhancing information retrieval across extended input sequences.
\end{itemize}

Our findings highlight pause tokens as an effective and computationally efficient mechanism for mitigating long-context deficiencies in LLMs.

\section{Related Work}

\subsection{Lost-in-the-Middle}

Large language models exhibit a U-shaped performance curve when processing long inputs, demonstrating a pronounced primacy and recency bias \citep{liu-etal-2024-lost}. That is, they allocate greater attention to the beginning and end of a sequence while neglecting the middle \citep{khandelwal-etal-2018-sharp, press-etal-2021-shortformer}. ~\citet{xiao2023efficient} further observed that models assign disproportionately high attention scores to initial tokens, even when these tokens lack semantic significance. This phenomenon extends to multi-document question-answering tasks—closely related to our evaluation framework—as well as key-value retrieval tasks. Building on these findings, our research investigates whether similar biases emerge in our specific context and examines their implications.

\subsection{Positional Encodings}

Altering attention through positional encoding is another promising approach to addressing LITM \citep{shaw2018selfattentionrelativepositionrepresentations}. \citet{su2023roformerenhancedtransformerrotary} proposed Rotary Position Embedding (RoPE), which encodes relative position information through a rotation matrix applied to token embeddings. RoPE allows for better extrapolation to longer sequences and has been widely adopted in recent language models \citep{deepseekai2024deepseekv3technicalreport, grattafiori2024llama3herdmodels}. \citet{zhang2024found} explores flaws with RoPE and seeks to address it using positional re-scaling. Modern implementations of RoPE, such as YaRN \citep{peng2023yarnefficientcontextwindow}, provide efficient scaling of context windows to extreme lengths (1M+ tokens). Positional encoding techniques show that attention recalibration can improve accuracy, while avoiding significant computation and fine-tuning. Rather than modifying positional embeddings, our work aims to enhance performance by redistributing attention through prompt editing and fine-tuning.

\subsection{Pause Tokens}

\citet{goyal2024thinkspeaktraininglanguage} introduced the concept of pause tokens in language model training. Their approach involves inserting learnable pause tokens during pretraining and finetuning, showing improvements on various Question-Answer tasks \cite{kwiatkowski-etal-2019-natural, talmor2019commonsenseqaquestionansweringchallenge, rajpurkar2016squad100000questionsmachine}. Expanding on this idea, \citet{rawte2024sorrycomeagainprompting} proposed the "Sorry, Come Again" (SCA) prompting technique, which integrates optimal paraphrasing with pause token injection. This method has been shown to effectively mitigate hallucinations in large language models, further underscoring the potential of pause tokens in enhancing model reliability and interpretability.

\section{Method}

\subsection{Token Placement Strategy}

We employ a systematic approach to injecting pause tokens across different experimental configurations. The most straightforward implementation involves inserting a special "<PAUSE>" token after each paragraph in the testing context. This token serves as a standardized pause marker, facilitating natural segmentation within the input sequence. By introducing these structured breaks, our approach enables the model to redistribute attention more effectively, ensuring comprehensive processing of information from all parts of the input.

\subsection{Experimental Configurations}

Our study explores various approaches for pause token insertion, as illustrated in \autoref{fig:pause_tuning}, using trials without input sequence modifications as the baseline for comparison. We evaluate the following techniques across 15 context depths and 3 trials for single needle tests and 15 randomized trials for multi-needle tests to identify the optimal method:
\begin{enumerate} 
    \item Standard Pause Tokens: Standard pause tokens are inserted after every paragraph in the input sequence. 
    \item Instruction-Augmented Pause Tokens: Pause tokens, including an explicit instruction to "stop and absorb the information [the model] has just read," are inserted after every paragraph in the input sequence. 
    \item Pre-Prompt Instruction with Standard Pause Tokens: A general instruction at the beginning of the prompt directs the model to stop and absorb information after every pause token, while standard pause tokens are inserted after every paragraph in the input sequence. 
    \item Fine-Tuning for Long Contexts: No modifications are made to the input sequence; instead, the model is fine-tuned for long-context comprehension using a one-shot prompt.
    \item Pause-Tuned Model with Standard Pause Tokens: Standard pause tokens are inserted after every paragraph in the input sequence, which is then processed by a model fine-tuned with standard pause tokens in long contexts.
\end{enumerate}

\noindent
These techniques aim to systematically determine the optimal method for enhancing the attention mechanism in long-context tasks.

Technique 2 and Technique 3 evaluate whether the model requires additional instructions or can naturally interpret pause tokens. Technique 4 serves as a control for Technique 5, assessing whether pause tokens are necessary for performance improvement or if long-context fine-tuning alone can achieve similar enhancements. Technique 5 employs pause-tuning, aligning the model with the pause token structure to enhance performance.

\begin{figure}
    \centering
    \includegraphics[width=1\linewidth]{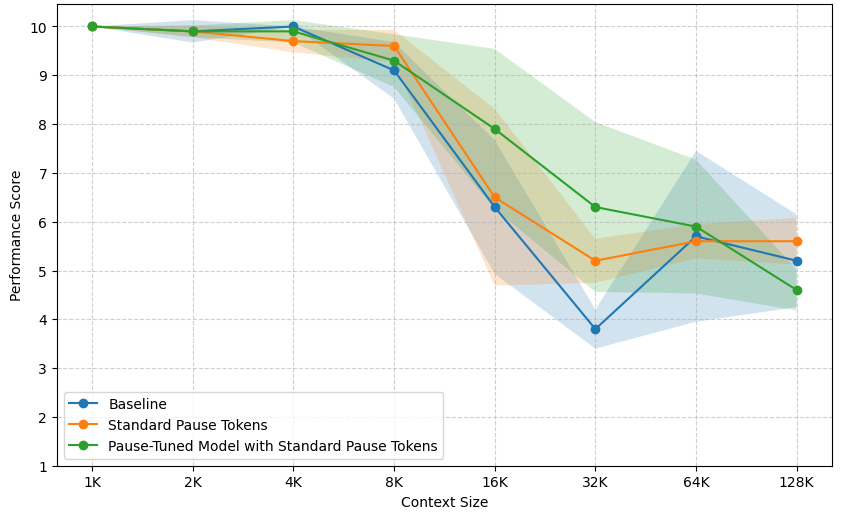}
    \caption{LLaMA 3.2 3B Instruct Performance Across Techniques. The score declines as context length increases. Pause-tuning consistently outperforms other methods, except at the 128K length.}
    \label{fig:llama_3.2_3b}
\end{figure}

\begin{figure}
    \centering
    \includegraphics[width=1\linewidth]{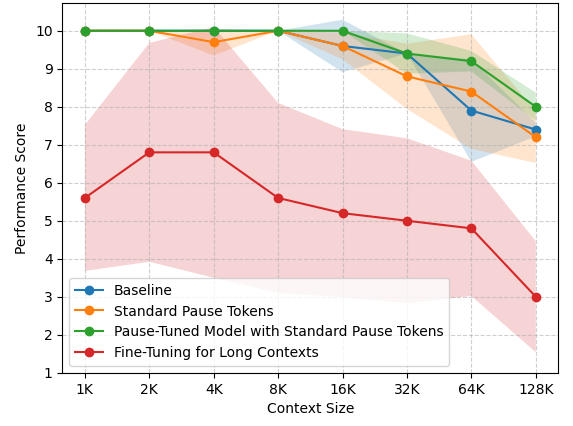}
    \caption{LLaMA 3.1 8B Instruct Performance Across Techniques. The score declines as context length increases. Pause-tuning consistently outperforms other methods at all context lengths.}
    \label{fig:llama_3.1_8b}
\end{figure}

\section{Experiments}

\subsection{Models}

We evaluate several widely used large language models (LLMs), including GPT-3.5 Turbo 0125 \citep{brown2020languagemodelsfewshotlearners}, GPT-4o Mini 2024-07-18 \citep{openai2024gpt4ocard}, LLaMA 3.2 1B Instruct, LLaMA 3.2 3B Instruct, and LLaMA 3.1 8B Instruct \citep{grattafiori2024llama3herdmodels}. For Technique 5, we employ pause-tuned versions of the LLaMA 3.2 3B Instruct and LLaMA 3.1 8B Instruct models.

\subsection{Evaluation}

We conduct our experiments using the Needle-in-a-Haystack evaluation framework \citep{needleinahaystack2023}. In this framework, a critical piece of information (the "needle") is embedded within a lengthy context (the "haystack"). Our evaluation spans various context lengths, ranging from 1K to 128K tokens, depending on each model's capacity for long contexts. Additionally, we assess both single-needle and multi-needle scenarios to examine the effectiveness of our method across different conditions. For multi-needle inputs, we embed three distinct needles within the context.

The results from each trial are assigned a score from 1 to 10 based on the success of the information retrieval attempt, as described in Appendix ~\ref{appendix_a}.

\begin{table*}[h!]
\centering
\small
\scalebox{0.85}
{
\begin{tabular}{@{}lcccccccc@{}}
\toprule
\textbf{Model}              & \textbf{1K} & \textbf{2K} & \textbf{4K} & \textbf{8K} & \textbf{16K} & \textbf{32K} & \textbf{64K} & \textbf{128K} \\ 
\midrule
\multicolumn{9}{c}{\textbf{Baseline}} \\ \midrule
GPT 3.5 & 10.00 \(\pm\) 0.00 & 10.00 \(\pm\) 0.00 & 10.00 \(\pm\) 0.00 & 10.00 \(\pm\) 0.00 & 9.27 \(\pm\) 0.60 & - & - & - \\
GPT 4o & 10.00 \(\pm\) 0.00 & 10.00 \(\pm\) 0.00 & 10.00 \(\pm\) 0.00 & 10.00 \(\pm\) 0.00 & 10.00 \(\pm\) 0.00 & 9.76 \(\pm\) 0.08 & 9.27 \(\pm\) 0.20 & 8.84 \(\pm\) 0.31 \\
LLaMA 3.2 1B & 7.67 \(\pm\) 0.93 & 6.64 \(\pm\) 1.86 & 4.11 \(\pm\) 1.89 & 3.67 \(\pm\) 0.81 & 1.4 \(\pm\) 0.38 & 1.38 \(\pm\) 0.69 & 1.04 \(\pm\) 0.08 & 1.00 \(\pm\) 0.00 \\
LLaMA 3.2 3B & 10.00 \(\pm\) 0.00 & 9.87 \(\pm\) 0.23 & 10.00 \(\pm\) 0.00 & 9.13 \(\pm\) 0.58 & 6.31 \(\pm\) 1.37 & 3.75 \(\pm\) 0.40 & 5.73 \(\pm\) 1.75 & 5.15 \(\pm\) 0.95 \\
LLaMA 3.1 8B & 10.00 \(\pm\) 0.00 & 10.00 \(\pm\) 0.00 & 10.00 \(\pm\) 0.00 & 10.00 \(\pm\) 0.00 & 9.60 \(\pm\) 0.69 & 9.40 \(\pm\) 0.00 & 7.89 \(\pm\) 1.35 & 7.40 \(\pm\) 0.18 \\ \midrule

\multicolumn{9}{c}{\textbf{Standard Pause Tokens (Technique 1)}} \\ \midrule
GPT 3.5 & 10.00 \(\pm\) 0.00 & 10.00 \(\pm\) 0.00 & 10.00 \(\pm\) 0.00 & 10.00 \(\pm\) 0.00 & 9.44 \(\pm\) 0.23 & - & - & - \\
GPT 4o & 10.00 \(\pm\) 0.00 & 10.00 \(\pm\) 0.00 & 10.00 \(\pm\) 0.00 & 10.00 \(\pm\) 0.00 & 10.00 \(\pm\) 0.00 & 10.00 \(\pm\) 0.00 & 9.71 \(\pm\) 0.08 & 9.47 \(\pm\) 0.00 \\
LLaMA 3.2 1B & 7.75 \(\pm\) 0.47 & 6.31 \(\pm\) 0.33 & 4.13 \(\pm\) 0.28 & 4.00 \(\pm\) 1.37 & 2.00 \(\pm\) 0.28 & 1.40 \(\pm\) 0.85 & 1.00 \(\pm\) 0.00 & 1.00 \(\pm\) 0.00 \\
LLaMA 3.2 3B & 10.00 \(\pm\) 0.00 & 9.93 \(\pm\) 0.12 & 9.67 \(\pm\) 0.23 & 9.62 \(\pm\) 0.33 & 6.51 \(\pm\) 1.81 & 5.24 \(\pm\) 0.45 & 5.60 \(\pm\) 0.35 & 5.64 \(\pm\) 0.48 \\
LLaMA 3.1 8B & 10.00 \(\pm\) 0.00 & 10.00 \(\pm\) 0.00 & 9.70 \(\pm\) 0.35 & 10.00 \(\pm\) 0.00 & 9.60 \(\pm\) 0.35 & 8.82 \(\pm\) 0.86 & 8.40 \(\pm\) 1.51 & 7.15 \(\pm\) 0.68 \\ \midrule

\multicolumn{9}{c}{\textbf{Instruction-Augmented Pause Tokens (Technique 2)}} \\ \midrule
GPT 3.5 & 10.00 $\pm$ 0.00 & 10.00 $\pm$ 0.00 & 10.00 $\pm$ 0.00 & 10.00 $\pm$ 0.00 & 9.44 $\pm$ 0.12 & - & - & - \\
GPT 4o & 10.00 $\pm$ 0.00 & 10.00 $\pm$ 0.00 & 10.00 $\pm$ 0.00 & 10.00 $\pm$ 0.00 & 10.00 $\pm$ 0.00 & 10.00 $\pm$ 0.00 & 10.00 $\pm$ 0.00 & 9.54 $\pm$ 0.12 \\
LLaMA 3.2 1B & 7.96 $\pm$ 0.66 & 6.60 $\pm$ 1.94 & 5.31 $\pm$ 2.86 & 3.14 $\pm$ 1.30 & 1.87 $\pm$ 0.83 & 1.04 $\pm$ 0.39 & 1.00 $\pm$ 0.00 & 1.00 $\pm$ 0.00 \\
LLaMA 3.2 3B & 10.00 $\pm$ 0.00 & 9.93 $\pm$ 0.12 & 9.93 $\pm$ 0.12 & 9.40 $\pm$ 0.35 & 6.35 $\pm$ 1.31 & 3.67 $\pm$ 0.37 & 5.36 $\pm$ 1.61 & 4.29 $\pm$ 1.49 \\
LLaMA 3.1 8B & 10.00 $\pm$ 0.00 & 10.00 $\pm$ 0.00 & 10.00 $\pm$ 0.00 & 10.00 $\pm$ 0.00 & 9.80 $\pm$ 0.35 & 8.80 $\pm$ 0.60 & 7.49 $\pm$ 0.20 & 6.27 $\pm$ 0.91 \\
 \midrule

\multicolumn{9}{c}{\textbf{Pre-Prompt Instruction with Standard Pause Tokens (Technique 3)}} \\ \midrule
GPT 3.5 & 10.00 $\pm$ 0.00 & 10.00 $\pm$ 0.00 & 10.00 $\pm$ 0.00 & 10.00 $\pm$ 0.00 & 9.44 $\pm$ 0.08 & - & - & - \\
GPT 4o & 10.00 $\pm$ 0.00 & 10.00 $\pm$ 0.00 & 10.00 $\pm$ 0.00 & 10.00 $\pm$ 0.00 & 10.00 $\pm$ 0.00 & 10.00 $\pm$ 0.00 & 9.71 $\pm$ 0.08 & 9.62 $\pm$ 0.08 \\
LLaMA 3.2 1B & 7.24 $\pm$ 0.60 & 7.62 $\pm$ 0.73 & 5.95 $\pm$ 2.31 & 3.73 $\pm$ 0.61 & 1.76 $\pm$ 0.33 & 1.69 $\pm$ 0.31 & 1.00 $\pm$ 0.00 & 1.00 $\pm$ 0.00 \\
LLaMA 3.2 3B & 9.80 $\pm$ 0.35 & 9.93 $\pm$ 0.12 & 10.00 $\pm$ 0.00 & 9.58 $\pm$ 0.40 & 6.02 $\pm$ 2.24 & 5.84 $\pm$ 0.77 & 4.47 $\pm$ 1.30 & 3.27 $\pm$ 1.01 \\
LLaMA 3.1 8B & 10.00 $\pm$ 0.00 & 10.00 $\pm$ 0.00 & 9.80 $\pm$ 0.35 & 9.80 $\pm$ 0.35 & 9.60 $\pm$ 0.35 & 9.20 $\pm$ 0.69 & 8.84 $\pm$ 0.53 & 7.15 $\pm$ 0.08 \\
 \midrule

\multicolumn{9}{c}{\textbf{Fine-Tuning for Long Contexts (Technique 4)}} \\ \midrule
LLaMA 3.1 8B & 5.60 \(\pm\) 1.92 & 6.80 \(\pm\) 2.88 & 6.80 \(\pm\) 3.30 & 5.60 \(\pm\) 2.50 & 5.20 \(\pm\) 2.21 & 5.00 \(\pm\) 2.17 & 4.80 \(\pm\) 1.78 & 3.00 \(\pm\) 1.46 \\ \midrule

\multicolumn{9}{c}{\textbf{Pause-Tuned Model with Standard Pause Tokens (Technique 5)}} \\ \midrule
LLaMA 3.2 3B & 10.00 \(\pm\) 0.00 & 9.93 \(\pm\) 0.12 & 9.87 \(\pm\) 0.23 & 9.29 \(\pm\) 0.54 & 7.91 \(\pm\) 1.64 & 6.29 \(\pm\) 1.74 & 5.89 \(\pm\) 1.37 & 4.56 \(\pm\) 0.42 \\
LLaMA 3.1 8B & 10.00 \(\pm\) 0.00 & 10.00 \(\pm\) 0.00 & 10.00 \(\pm\) 0.00 & 10.00 \(\pm\) 0.00 & 10.00 \(\pm\) 0.00 & 9.44 \(\pm\) 0.53 & 9.16 \(\pm\) 0.27 & 7.98 \(\pm\) 0.37 \\ \midrule

\multicolumn{9}{c}{\textbf{Pause-Tuning (\% Change Over Baseline)}} \\ \midrule
LLaMA 3.2 3B & 0.00 & 0.61 & -1.30 & 1.75 & 25.36 & 67.73 & 2.79 & -11.46 \\
LLaMA 3.1 8B  & 0.00 & 0.00 & 0.00 & 0.00 & 4.17 & 0.43 & 16.10 & 7.84 \\
\bottomrule

\end{tabular}
}
\caption{Results of the single needle task, presented as mean $\pm$ standard deviation, comparing the baseline, pause token methods, and pause-tuning. Token counts are denoted as 1K = 1,000 tokens, 2K = 2,000 tokens, etc.}
\label{tab:combined_results}
\end{table*}

\subsection{Datasets}

We use the Deep Essays Dataset \citep{willian_oliveira_gibin_mann_acharya_2024} and the DAIGT Gemini-Pro 8.5K Essays dataset \citep{demir_essays_dataset} for fine-tuning and a collection of Paul Graham's essays \citep{graham_essays} to create the haystack for testing.

\subsection{Pause-Tuning}

We fine-tune two LLaMA models for pause tokens in long contexts. These models were selected for their ability to handle sequences of up to 128K tokens and their strong performance in instructional tasks.

To construct an appropriate fine-tuning dataset, we concatenate multiple shorter essays and systematically inject pause tokens until the target context length is reached. Additionally, we embed a randomly selected piece of information—a "needle"—within this extended context "haystack" to assess retrieval capabilities. The models are trained using LoRA \citep{lora_hu_shen} and Unsloth AI \citep{unsloth}. The hyperparameters for training are in Appendix ~\ref{appendix:para}. The training prompt adopts a one-shot format, consisting of four key components: an instruction outlining the purpose of the fine-tuning, a long-context input with injected pauses that contains the needle, an example user query that depends on retrieving the embedded information, and a response that reproduces the needle verbatim as it was originally inserted into the essay.


\section{Results}

The results for the baseline and each technique using a single needle are presented in \autoref{tab:combined_results}, while the results for multiple needles can be found in Appendix~\ref{appendix:multi}. A visualization of the performance scores can be found in ~\autoref{fig:llama_3.2_3b} and~\autoref{fig:llama_3.1_8b}. A significant improvement is observed in the pause-tuned models compared to both the baseline and the other techniques. While Technique 3 proves highly effective (11.44\% improvement)  on the LLaMA 3.2 1B model, as seen in \autoref{tab:percent_change}, its impact is less pronounced on the other models. In contrast, Technique 5 demonstrates substantial improvements, with 10.61\% and 3.57\% increases, across both models, supporting our hypothesis that combining fine-tuning with pause tokens yields the best results. This outcome aligns with our intuition that training the model with a consistent structural approach enhances performance. 

Notably, the LLaMA 3.1 8B Instruct model, evaluated across all techniques, exhibits a 16.10\% improvement over the baseline at 64K tokens and a 7.84\% improvement at 128K tokens using the pause-tuning technique. These results highlight the effectiveness of integrating pause tokens within the fine-tuning process. In contrast, Techniques 1 and 4, which involve using pause tokens without fine-tuning and fine-tuning without pause tokens, respectively, prove highly ineffective. These findings demonstrate that neither pause tokens nor fine-tuning alone significantly enhance performance. However, when the model is explicitly trained to recognize and utilize the pause token, its ability to process long-context inputs improves remarkably.


\begin{table}[h!]
\centering
\small
\scalebox{1}{
\begin{tabular}{@{}lcccccccc@{}}
\toprule
\textbf{Technique}              & \textbf{1} & \textbf{2} & \textbf{3} & \textbf{4} & \textbf{5} \\ \midrule
GPT 3.5          &0.37&0.37&0.37&---&---          \\
GPT 4o         &1.68&2.15&1.87&---&---          \\
LLaMA 3.2 1B       &2.53&3.75&11.44&---&---            \\
LLaMA 3.2 3B       &1.03&-1.70&-1.72&---&10.61          \\
LLaMA 3.1 8B       &-1.67&-2.60&0.14&-42.88&3.57          \\
\bottomrule
\end{tabular}
}
\caption{Percent Change for each technique compared to baseline, averaged across all context lengths.}
\label{tab:percent_change}
\end{table}

\section{Attention Analysis}

\begin{figure}[h]
    
    \includegraphics[width=1\linewidth]{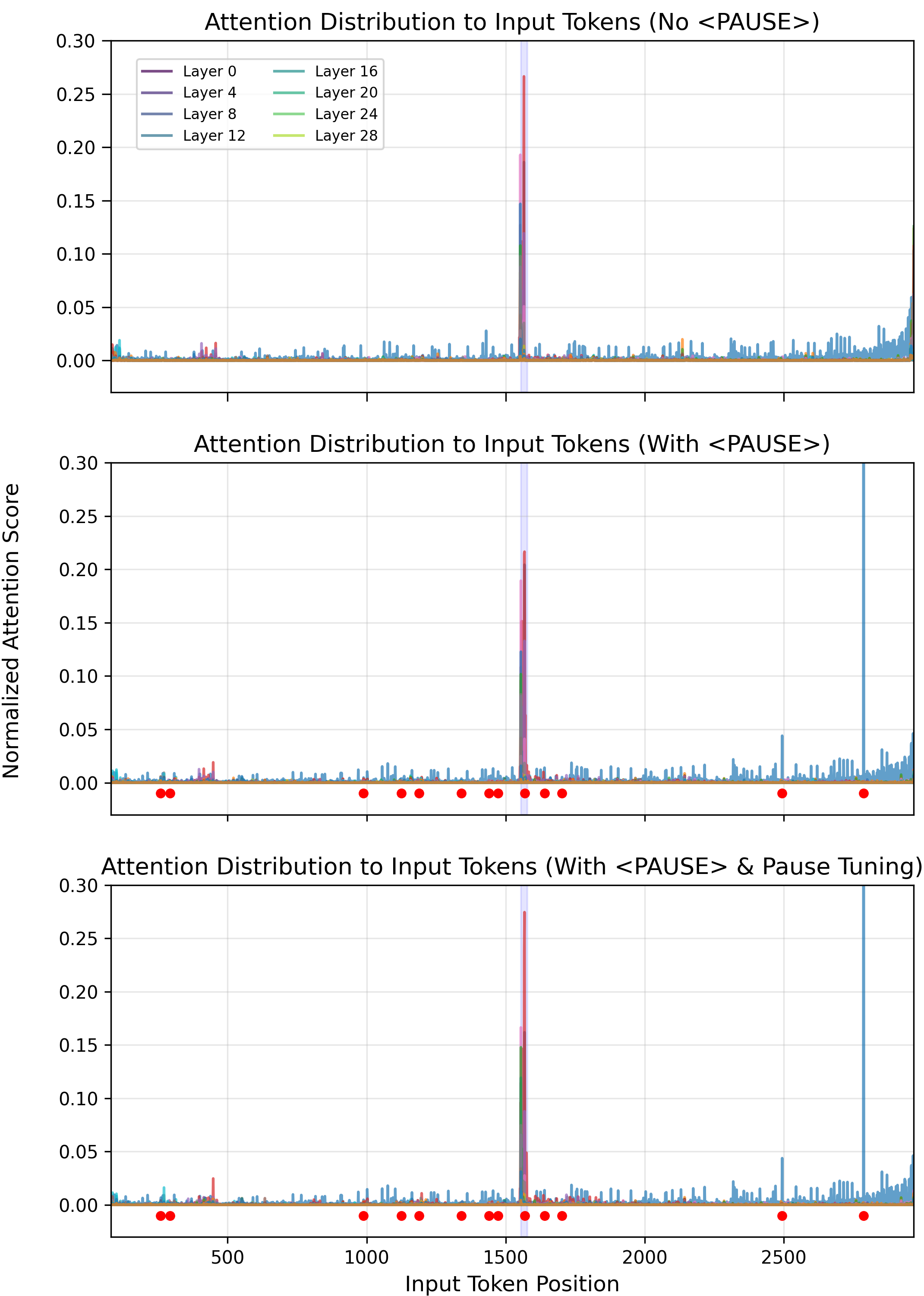}
    \caption{Attention distribution with normalized attention scores across different layers when generating the first answer token for baseline, standard pause token insertion, and pause-tuned with pause tokens inserted. The purple highlight indicates the position of the needle. The red dots indicate inserted pause tokens.}
    \label{fig:baseline_attention}
\end{figure}

\autoref{fig:baseline_attention} depicts the scaled attention distribution across different layers when generating the first answer token for the baseline and Techniques 1 and 5. Due to computational constraints, these results are for 3000 token input sequences. However, the observed retrieval improvements for sequences up to 128K tokens suggest that this pattern likely extends to longer contexts. 


The insertion of pause tokens significantly transforms the attention distribution. We observe distinct attention spikes at the locations of several pause tokens. This distinction is especially present surrounding the needle and in the latter half of the context for both the standard pause token insertion and the pause-tuned model. These findings indicate that pause tokens serve as anchors that interrupt the attention decay over long sequences, prompting the model to engage with all sections more thoroughly. The model likely treats the paragraphs separated by pause tokens as distinct sections, refreshing attention and decreasing the likelihood of forgetting important information. This structured attention recalibration may explain the improved retrieval performance observed in longer contexts.

\section{Conclusion}

In this paper, we introduce pause-tuning as an effective and lightweight solution to the LITM problem, addressing long-context comprehension challenges in LLMs. By strategically injecting pause tokens into input sequences, we enhance attention redistribution, enabling models to retrieve information more effectively across extensive contexts. Our experiments demonstrate significant improvements in retrieval performance over baseline models in the Needle-in-a-Haystack benchmark. The results confirm that pause-tuning consistently enhances long-context retention across various input lengths. These findings underscore the potential of pause-tuning as a practical technique for mitigating LITM issues, paving the way for more robust long-context processing in future LLM advancements.


\section*{Limitation}

One limitation of our evaluation framework is its focus on a specific retrieval method, where a fact is embedded within a long context. While our results suggest that the attention shifts induced by pause-tuning enhance overall comprehension, we cannot generalize this improvement to all long-context use cases. In many scenarios, pure retrieval is insufficient; instead, effective reasoning and understanding of interrelated information within long contexts are crucial.

Another limitation is that our study evaluates the pause-tuning technique only on relatively small models (<10B parameters). Whether these findings extend to large-scale models remains an open question and requires further investigation.

\bibliography{anthology, main}

\begin{thebibliography}{28}
\providecommand{\natexlab}[1]{#1}

\bibitem[{Brown et~al.(2020)Brown, Mann, Ryder, Subbiah, Kaplan, Dhariwal, Neelakantan, Shyam, Sastry, Askell, Agarwal, Herbert-Voss, Krueger, Henighan, Child, Ramesh, Ziegler, Wu, and Amodei}]{brown2020languagemodelsfewshotlearners}
Tom~B. Brown, Benjamin Mann, Nick Ryder, Melanie Subbiah, Jared Kaplan, Prafulla Dhariwal, Arvind Neelakantan, Pranav Shyam, Girish Sastry, Amanda Askell, Sandhini Agarwal, Ariel Herbert-Voss, Gretchen Krueger, Tom Henighan, Rewon Child, Aditya Ramesh, Daniel~M. Ziegler, Jeffrey Wu, and Clemens Winter ...~Dario Amodei. 2020.
\newblock \href {https://arxiv.org/abs/2005.14165} {Language models are few-shot learners}.
\newblock \emph{Preprint}, arXiv:2005.14165.

\bibitem[{DeepSeek-AI et~al.(2024)DeepSeek-AI, Liu, Feng, Xue, Wang, Wu, Lu, Zhao, Deng, Zhang, Ruan, Dai, Guo, Yang, Chen, Ji, Li, Lin, Dai, and Pan}]{deepseekai2024deepseekv3technicalreport}
DeepSeek-AI, Aixin Liu, Bei Feng, Bing Xue, Bingxuan Wang, Bochao Wu, Chengda Lu, Chenggang Zhao, Chengqi Deng, Chenyu Zhang, Chong Ruan, Damai Dai, Daya Guo, Dejian Yang, Deli Chen, Dongjie Ji, Erhang Li, Fangyun Lin, Fucong Dai, and Fuli Luo ... H.~Zizheng Pan. 2024.
\newblock \href {https://arxiv.org/abs/2412.19437} {Deepseek-v3 technical report}.
\newblock \emph{Preprint}, arXiv:2412.19437.

\bibitem[{Demir(2024)}]{demir_essays_dataset}
Ertuğrul Demir. 2024.
\newblock \href {https://www.kaggle.com/datasets/datafan07/daigt-gemini-pro-8-5k-essays} {Daigt gemini-pro 8.5k essays}.

\bibitem[{Gibin and Acharya(2024)}]{willian_oliveira_gibin_mann_acharya_2024}
Willian~Oliveira Gibin and Mann Acharya. 2024.
\newblock \href {https://doi.org/10.34740/KAGGLE/DSV/7607799} {Deep essays dataset}.

\bibitem[{Goyal et~al.(2024)Goyal, Ji, Rawat, Menon, Kumar, and Nagarajan}]{goyal2024thinkspeaktraininglanguage}
Sachin Goyal, Ziwei Ji, Ankit~Singh Rawat, Aditya~Krishna Menon, Sanjiv Kumar, and Vaishnavh Nagarajan. 2024.
\newblock \href {https://arxiv.org/abs/2310.02226} {Think before you speak: Training language models with pause tokens}.
\newblock \emph{Preprint}, arXiv:2310.02226.

\bibitem[{Graham(2001/2023)}]{graham_essays}
Paul Graham. 2001/2023.
\newblock Essays.
\newblock \url{http://www.paulgraham.com/articles.html}.
\newblock A collection of essays published online between 2001 and 2023.

\bibitem[{Grattafiori et~al.(2024)Grattafiori, Dubey, Jauhri, Pandey, Kadian, Al-Dahle, Letman, Mathur, Schelten, Vaughan, Yang, Fan, Goyal, Hartshorn, Yang, Mitra, Sravankumar, Korenev, and Ma}]{grattafiori2024llama3herdmodels}
Aaron Grattafiori, Abhimanyu Dubey, Abhinav Jauhri, Abhinav Pandey, Abhishek Kadian, Ahmad Al-Dahle, Aiesha Letman, Akhil Mathur, Alan Schelten, Alex Vaughan, Amy Yang, Angela Fan, Anirudh Goyal, Anthony Hartshorn, Aobo Yang, Archi Mitra, Archie Sravankumar, Artem Korenev, and Arthur Hinsvark ...~Zhiyu Ma. 2024.
\newblock \href {https://arxiv.org/abs/2407.21783} {The llama 3 herd of models}.
\newblock \emph{Preprint}, arXiv:2407.21783.

\bibitem[{Han et~al.(2023)Han, Han, and team}]{unsloth}
Daniel Han, Michael Han, and Unsloth team. 2023.
\newblock \href {http://github.com/unslothai/unsloth} {Unsloth}.

\bibitem[{He et~al.(2024)He, Pan, Dong, Song, Liu, Sun, Liang, Wang, Zhang, and Zhang}]{he2024lostmiddlemasteringlongcontext}
Junqing He, Kunhao Pan, Xiaoqun Dong, Zhuoyang Song, Yibo Liu, Qianguo Sun, Yuxin Liang, Hao Wang, Enming Zhang, and Jiaxing Zhang. 2024.
\newblock \href {https://arxiv.org/abs/2311.09198} {Never lost in the middle: Mastering long-context question answering with position-agnostic decompositional training}.
\newblock \emph{Preprint}, arXiv:2311.09198.

\bibitem[{Hu et~al.(2021)Hu, Shen, Wallis, Allen{-}Zhu, Li, Wang, and Chen}]{lora_hu_shen}
Edward~J. Hu, Yelong Shen, Phillip Wallis, Zeyuan Allen{-}Zhu, Yuanzhi Li, Shean Wang, and Weizhu Chen. 2021.
\newblock \href {https://arxiv.org/abs/2106.09685} {Lora: Low-rank adaptation of large language models}.
\newblock \emph{CoRR}, abs/2106.09685.

\bibitem[{Kamradt(2023)}]{needleinahaystack2023}
Greg Kamradt. 2023.
\newblock Needleinahaystack.
\newblock \url{https://github.com/gkamradt/LLMTest_NeedleInAHaystack}.

\bibitem[{Khandelwal et~al.(2018)Khandelwal, He, Qi, and Jurafsky}]{khandelwal-etal-2018-sharp}
Urvashi Khandelwal, He~He, Peng Qi, and Dan Jurafsky. 2018.
\newblock \href {https://doi.org/10.18653/v1/P18-1027} {Sharp nearby, fuzzy far away: How neural language models use context}.
\newblock In \emph{Proceedings of the 56th Annual Meeting of the Association for Computational Linguistics (Volume 1: Long Papers)}, pages 284--294, Melbourne, Australia. Association for Computational Linguistics.

\bibitem[{Kwiatkowski et~al.(2019)Kwiatkowski, Palomaki, Redfield, Collins, Parikh, Alberti, Epstein, Polosukhin, Devlin, Lee, Toutanova, Jones, Kelcey, Chang, Dai, Uszkoreit, Le, and Petrov}]{kwiatkowski-etal-2019-natural}
Tom Kwiatkowski, Jennimaria Palomaki, Olivia Redfield, Michael Collins, Ankur Parikh, Chris Alberti, Danielle Epstein, Illia Polosukhin, Jacob Devlin, Kenton Lee, Kristina Toutanova, Llion Jones, Matthew Kelcey, Ming-Wei Chang, Andrew~M. Dai, Jakob Uszkoreit, Quoc Le, and Slav Petrov. 2019.
\newblock \href {https://doi.org/10.1162/tacl_a_00276} {Natural questions: A benchmark for question answering research}.
\newblock \emph{Transactions of the Association for Computational Linguistics}, 7:452--466.

\bibitem[{Liu and Abbeel(2023)}]{liu2023blockwiseparalleltransformerlarge}
Hao Liu and Pieter Abbeel. 2023.
\newblock \href {https://arxiv.org/abs/2305.19370} {Blockwise parallel transformer for large context models}.
\newblock \emph{Preprint}, arXiv:2305.19370.

\bibitem[{Liu et~al.(2024)Liu, Lin, Hewitt, Paranjape, Bevilacqua, Petroni, and Liang}]{liu-etal-2024-lost}
Nelson~F. Liu, Kevin Lin, John Hewitt, Ashwin Paranjape, Michele Bevilacqua, Fabio Petroni, and Percy Liang. 2024.
\newblock \href {https://doi.org/10.1162/tacl_a_00638} {Lost in the middle: How language models use long contexts}.
\newblock \emph{Transactions of the Association for Computational Linguistics}, 12:157--173.

\bibitem[{Minaee et~al.(2024)Minaee, Mikolov, Nikzad, Chenaghlu, Socher, Amatriain, and Gao}]{minaee2024largelanguagemodelssurvey}
Shervin Minaee, Tomas Mikolov, Narjes Nikzad, Meysam Chenaghlu, Richard Socher, Xavier Amatriain, and Jianfeng Gao. 2024.
\newblock \href {https://arxiv.org/abs/2402.06196} {Large language models: A survey}.
\newblock \emph{Preprint}, arXiv:2402.06196.

\bibitem[{Nelson et~al.(2024)Nelson, Kollias, Das, Chaudhury, and Dan}]{nelson2024needlehaystackmemorybased}
Elliot Nelson, Georgios Kollias, Payel Das, Subhajit Chaudhury, and Soham Dan. 2024.
\newblock \href {https://arxiv.org/abs/2407.01437} {Needle in the haystack for memory based large language models}.
\newblock \emph{Preprint}, arXiv:2407.01437.

\bibitem[{OpenAI et~al.(2024)OpenAI, :, Hurst, Lerer, Goucher, Perelman, Ramesh, Clark, Ostrow, Welihinda, Hayes, Radford, Mądry, Baker-Whitcomb, Beutel, Borzunov, Carney, Chow, Kirillov, Nichol, and Malkov}]{openai2024gpt4ocard}
OpenAI, :, Aaron Hurst, Adam Lerer, Adam~P. Goucher, Adam Perelman, Aditya Ramesh, Aidan Clark, AJ~Ostrow, Akila Welihinda, Alan Hayes, Alec Radford, Aleksander Mądry, Alex Baker-Whitcomb, Alex Beutel, Alex Borzunov, Alex Carney, Alex Chow, Alex Kirillov, Alex Nichol, and Alex Paino ...~Yury Malkov. 2024.
\newblock \href {https://arxiv.org/abs/2410.21276} {Gpt-4o system card}.
\newblock \emph{Preprint}, arXiv:2410.21276.

\bibitem[{Peng et~al.(2023)Peng, Quesnelle, Fan, and Shippole}]{peng2023yarnefficientcontextwindow}
Bowen Peng, Jeffrey Quesnelle, Honglu Fan, and Enrico Shippole. 2023.
\newblock \href {https://arxiv.org/abs/2309.00071} {Yarn: Efficient context window extension of large language models}.
\newblock \emph{Preprint}, arXiv:2309.00071.

\bibitem[{Press et~al.(2021)Press, Smith, and Lewis}]{press-etal-2021-shortformer}
Ofir Press, Noah~A. Smith, and Mike Lewis. 2021.
\newblock \href {https://doi.org/10.18653/v1/2021.acl-long.427} {Shortformer: Better language modeling using shorter inputs}.
\newblock In \emph{Proceedings of the 59th Annual Meeting of the Association for Computational Linguistics and the 11th International Joint Conference on Natural Language Processing (Volume 1: Long Papers)}, pages 5493--5505, Online. Association for Computational Linguistics.

\bibitem[{Rajpurkar et~al.(2016)Rajpurkar, Zhang, Lopyrev, and Liang}]{rajpurkar2016squad100000questionsmachine}
Pranav Rajpurkar, Jian Zhang, Konstantin Lopyrev, and Percy Liang. 2016.
\newblock \href {https://arxiv.org/abs/1606.05250} {Squad: 100,000+ questions for machine comprehension of text}.
\newblock \emph{Preprint}, arXiv:1606.05250.

\bibitem[{Rawte et~al.(2024)Rawte, Tonmoy, Zaman, Priya, Chadha, Sheth, and Das}]{rawte2024sorrycomeagainprompting}
Vipula Rawte, S.~M Towhidul~Islam Tonmoy, S~M~Mehedi Zaman, Prachi Priya, Aman Chadha, Amit~P. Sheth, and Amitava Das. 2024.
\newblock \href {https://arxiv.org/abs/2403.18976} {"sorry, come again?" prompting -- enhancing comprehension and diminishing hallucination with [pause]-injected optimal paraphrasing}.
\newblock \emph{Preprint}, arXiv:2403.18976.

\bibitem[{Shaw et~al.(2018)Shaw, Uszkoreit, and Vaswani}]{shaw2018selfattentionrelativepositionrepresentations}
Peter Shaw, Jakob Uszkoreit, and Ashish Vaswani. 2018.
\newblock \href {https://arxiv.org/abs/1803.02155} {Self-attention with relative position representations}.
\newblock \emph{Preprint}, arXiv:1803.02155.

\bibitem[{Su et~al.(2023)Su, Lu, Pan, Murtadha, Wen, and Liu}]{su2023roformerenhancedtransformerrotary}
Jianlin Su, Yu~Lu, Shengfeng Pan, Ahmed Murtadha, Bo~Wen, and Yunfeng Liu. 2023.
\newblock \href {https://arxiv.org/abs/2104.09864} {Roformer: Enhanced transformer with rotary position embedding}.
\newblock \emph{Preprint}, arXiv:2104.09864.

\bibitem[{Talmor et~al.(2019)Talmor, Herzig, Lourie, and Berant}]{talmor2019commonsenseqaquestionansweringchallenge}
Alon Talmor, Jonathan Herzig, Nicholas Lourie, and Jonathan Berant. 2019.
\newblock \href {https://arxiv.org/abs/1811.00937} {Commonsenseqa: A question answering challenge targeting commonsense knowledge}.
\newblock \emph{Preprint}, arXiv:1811.00937.

\bibitem[{Tworkowski et~al.(2023)Tworkowski, Staniszewski, Pacek, Wu, Michalewski, and Miłoś}]{tworkowski2023focusedtransformercontrastivetraining}
Szymon Tworkowski, Konrad Staniszewski, Mikołaj Pacek, Yuhuai Wu, Henryk Michalewski, and Piotr Miłoś. 2023.
\newblock \href {https://arxiv.org/abs/2307.03170} {Focused transformer: Contrastive training for context scaling}.
\newblock \emph{Preprint}, arXiv:2307.03170.

\bibitem[{Xiao et~al.(2023)Xiao, Tian, Chen, Han, and Lewis}]{xiao2023efficient}
Guangxuan Xiao, Yuandong Tian, Beidi Chen, Song Han, and Mike Lewis. 2023.
\newblock Efficient streaming language models with attention sinks.
\newblock \emph{arXiv preprint arXiv:2309.17453}.

\bibitem[{Zhang et~al.(2024)Zhang, Chen, Liu, Yao, Ruwase, Chen, Wu, and Wang}]{zhang2024found}
Zhenyu Zhang, Runjin Chen, Shiwei Liu, Zhewei Yao, Olatunji Ruwase, Beidi Chen, Xiaoxia Wu, and Zhangyang Wang. 2024.
\newblock \href {https://arxiv.org/abs/2403.04797} {Found in the middle: How language models use long contexts better via plug-and-play positional encoding}.
\newblock \emph{arXiv preprint arXiv:2403.04797}.

\end{thebibliography}

\clearpage
\appendix
\section{Needle-in-a-Haystack Evaluation}
\label{appendix_a}

To score the retrievals for the needle-in-a-haystack test, the following framework was used, with relevance assessed by the model: \\\\
Score 1: The answer is completely unrelated to the reference. \\
                Score 3: The answer has minor relevance but does not align with the reference.\\
                Score 5: The answer has moderate relevance but contains inaccuracies.\\
                Score 7: The answer aligns with the reference but has minor omissions.\\
                Score 10: The answer is completely accurate and aligns perfectly with the reference.

\section{Prompt Formatting}

\begin{tcolorbox}[colframe=green!50!black, colback=gray!10, title=Prompt without <PAUSE> tokens]
Below is an instruction that describes a task, paired with a context that provides further information. An input will request information from the context. Write a response that appropriately completes the request.\\

\#\#\#Instruction:\\\\
You are a helpful assistant that will be provided a context which the user wants to ask a question about, your job is to answer the question with only statements provided in the context and nothing else.\\

\#\#\#Context:\\\\
\{context\}\\

\#\#\#Input:\\\\
\{input\}
\end{tcolorbox}

\begin{tcolorbox}[colframe=green!50!black, colback=gray!10, title=Prompt with <PAUSE> tokens]
Below is an instruction that describes a task, paired with a context that provides further information. An input will request information from the context. Write a response that appropriately completes the request.\\

\#\#\#Instruction:\\\\
You are a helpful assistant that will be provided a context which the user wants to ask a question about, the context has <PAUSE> tokens that tell you when to take a pause to comprehend the context before continuing, your job is to answer the question with only statements provided in the context and nothing else.\\

\#\#\#Context:\\\\
\{context\}\\

\#\#\#Input:\\\\
\{input\}
\end{tcolorbox}

\clearpage

\section{Hyperparameters}
\label{appendix:para}

The parameters in \autoref{tab:parameters} were used as part of the fine-tuning process.

\begin{table*}[h]
    \centering
    \scalebox{0.9}{
    \begin{tabular}{|l|l|}
        \hline
        \multicolumn{2}{|c|}{\textbf{LoRA Parameters}} \\
        \hline
        Rank (r) & 16 \\
        Alpha & 16 \\
        Dropout & 0 \\
        Target Modules & ["q\_proj", "k\_proj", "v\_proj", "o\_proj", "gate\_proj", "up\_proj", "down\_proj"] \\
        \hline
        \multicolumn{2}{|c|}{\textbf{Training Parameters}} \\
        \hline
        Batch Size & 2 \\
        Gradient Accumulation Steps & 4 \\
        Learning Rate & 2e-4 \\
        Weight Decay & 0.01 \\
        Warmup Steps & 6 \\
        Steps & 60 \\
        LR Scheduler & linear \\
        Optimizer & adamw\_8bit \\
        \hline
        \multicolumn{2}{|c|}{\textbf{Other}} \\
        \hline
        Mixed Precision & fp16 \\
        Quantization Bits & 4 bit \\
        \hline
    \end{tabular}}
    \caption{Training Hyperparameters}
    \label{tab:parameters}
\end{table*}
\section{Multi-Needle Results}
\label{appendix:multi}

While the results in the main paper utilize a single needle, we also conduct tests with three needles in the haystack. The results are reported in \autoref{tab:combined_results_2}.

\begin{table*}[h!]
\centering
\small
\scalebox{0.85}
{
\begin{tabular}{@{}lcccccccc@{}}
\toprule
\textbf{Model} & \textbf{1K} & \textbf{2K} & \textbf{4K} & \textbf{8K} & \textbf{16K} & \textbf{32K} & \textbf{64K} & \textbf{128K} \\ \midrule
\multicolumn{9}{c}{\textbf{Baseline}} \\ \midrule
GPT 3.5 & 10.00 $\pm$ 0.00 & 9.60 $\pm$ 1.06 & 9.87 $\pm$ 0.52 & 8.00 $\pm$ 1.73 & 8.00 $\pm$ 1.36 & - & - & - \\
GPT 4o & 9.60 $\pm$ 1.06 & 10.00 $\pm$ 0.00 & 9.27 $\pm$ 1.55 & 8.80 $\pm$ 2.48 & 8.87 $\pm$ 2.26 & 7.20 $\pm$ 2.65 & 8.53 $\pm$ 2.64 & 7.60 $\pm$ 1.68 \\
LLaMA 3.2 1B & 6.40 $\pm$ 4.56 & 8.60 $\pm$ 3.18 & 7.00 $\pm$ 4.39 & 1.00 $\pm$ 0.00 & 3.60 $\pm$ 4.06 & 1.00 $\pm$ 0.00 & 2.20 $\pm$ 3.17 & 1.00 $\pm$ 0.00 \\
LLaMA 3.2 3B & 7.20 $\pm$ 3.10 & 6.20 $\pm$ 2.40 & 5.20 $\pm$ 2.21 & 3.60 $\pm$ 1.92 & 4.20 $\pm$ 2.11 & 3.40 $\pm$ 1.68 & 3.40 $\pm$ 2.32 & 2.20 $\pm$ 1.52 \\
LLaMA 3.1 8B & 8.07 $\pm$ 2.69 & 7.47 $\pm$ 1.92 & 6.80 $\pm$ 1.78 & 6.40 $\pm$ 1.68 & 4.40 $\pm$ 2.10 & 4.27 $\pm$ 1.55 & 4.27 $\pm$ 0.80 & 4.20 $\pm$ 0.77 \\
 \midrule

\multicolumn{9}{c}{\textbf{Standard Pause Tokens (Technique 1)}} \\ \midrule
GPT 3.5 & 10.00 $\pm$ 0.00 & 9.40 $\pm$ 1.24 & 9.93 $\pm$ 0.26 & 8.67 $\pm$ 2.17 & 7.67 $\pm$ 1.68 & - & - & - \\
GPT 4o & 9.20 $\pm$ 1.37 & 9.80 $\pm$ 0.77 & 9.60 $\pm$ 1.06 & 9.00 $\pm$ 1.46 & 8.47 $\pm$ 1.78 & 8.00 $\pm$ 2.45 & 7.80 $\pm$ 2.40 & 8.40 $\pm$ 1.92 \\
LLaMA 3.2 1B & 8.20 $\pm$ 3.73 & 9.40 $\pm$ 2.32 & 9.40 $\pm$ 2.32 & 6.20 $\pm$ 4.46 & 5.00 $\pm$ 4.49 & 3.40 $\pm$ 4.12 & 2.80 $\pm$ 3.72 & 1.00 $\pm$ 0.00 \\
LLaMA 3.2 3B & 8.00 $\pm$ 2.17 & 7.40 $\pm$ 1.92 & 6.13 $\pm$ 2.36 & 5.53 $\pm$ 1.85 & 4.80 $\pm$ 1.37 & 4.93 $\pm$ 1.75 & 4.27 $\pm$ 0.80 & 3.13 $\pm$ 2.72 \\
LLaMA 3.1 8B & 8.00 $\pm$ 2.70 & 7.40 $\pm$ 1.55 & 6.67 $\pm$ 2.50 & 7.00 $\pm$ 1.96 & 5.20 $\pm$ 1.52 & 5.00 $\pm$ 1.46 & 4.47 $\pm$ 1.55 & 4.20 $\pm$ 1.37 \\
 \midrule

\multicolumn{9}{c}{\textbf{Instruction-Augmented Pause Tokens (Technique 2)}} \\ \midrule
GPT 3.5 & 10.00 $\pm$ 0.00 & 10.00 $\pm$ 0.00 & 9.87 $\pm$ 0.52 & 8.53 $\pm$ 1.46 & 8.20 $\pm$ 1.37 & - & - & - \\
GPT 4o & 10.00 $\pm$ 0.00 & 9.00 $\pm$ 1.06 & 8.60 $\pm$ 1.92 & 9.40 $\pm$ 1.24 & 8.00 $\pm$ 2.45 & 7.00 $\pm$ 3.21 & 7.67 $\pm$ 2.32 & 7.00 $\pm$ 2.27 \\
LLaMA 3.2 1B & 3.00 $\pm$ 1.85 & 2.20 $\pm$ 1.90 & 2.80 $\pm$ 2.73 & 2.20 $\pm$ 2.21 & 1.40 $\pm$ 1.06 & 1.80 $\pm$ 2.40 & 1.60 $\pm$ 1.68 & 1.00 $\pm$ 0.00 \\
LLaMA 3.2 3B & 7.13 $\pm$ 2.59 & 5.67 $\pm$ 2.72 & 5.33 $\pm$ 1.29 & 4.27 $\pm$ 1.39 & 5.00 $\pm$ 2.07 & 4.07 $\pm$ 1.16 & 3.67 $\pm$ 1.11 & 3.33 $\pm$ 2.47 \\
LLaMA 3.1 8B & 9.00 $\pm$ 1.46 & 7.67 $\pm$ 1.54 & 7.47 $\pm$ 1.41 & 7.07 $\pm$ 0.96 & 6.73 $\pm$ 0.70 & 6.86 $\pm$ 0.52 & 6.60 $\pm$ 1.55 & 5.71 $\pm$ 1.68 \\
\midrule

\multicolumn{9}{c}{\textbf{Pre-Prompt Instruction with Standard Pause Tokens (Technique 3)}} \\ \midrule
GPT 3.5 & 10.00 \(\pm\) 0.00 & 9.60 \(\pm\) 1.55 & 9.60 \(\pm\) 1.06 & 8.93 \(\pm\) 1.44 & 7.87 \(\pm\) 2.10 & - & - & - \\
GPT 4o & 9.80 \(\pm\) 0.77 & 10.00 \(\pm\) 0.00 & 9.20 \(\pm\) 1.78 & 9.40 \(\pm\) 1.24 & 8.80 \(\pm\) 1.52 & 8.20 \(\pm\) 1.90 & 7.00 \(\pm\) 2.27 & 7.20 \(\pm\) 1.37 \\
LLaMA 3.2 1B & 2.40 \(\pm\) 1.55 & 2.20 \(\pm\) 1.52 & 2.40 \(\pm\) 1.55 & 2.80 \(\pm\) 1.90 & 1.40 \(\pm\) 1.06 & 1.60 \(\pm\) 1.68 & 1.40 \(\pm\) 1.06 & 1.00 \(\pm\) 0.00 \\
LLaMA 3.2 3B & 7.20 \(\pm\) 2.65 & 7.00 \(\pm\) 2.27 & 6.00 \(\pm\) 1.46 & 4.40 \(\pm\) 1.92 & 4.60 \(\pm\) 2.32 & 4.00 \(\pm\) 1.60 & 4.20 \(\pm\) 1.37 & 3.00 \(\pm\) 1.46 \\
LLaMA 3.1 8B & 8.20 \(\pm\) 1.52 & 7.40 \(\pm\) 1.95 & 7.60 \(\pm\) 1.24 & 7.07 \(\pm\) 1.22 & 7.27 \(\pm\) 0.80 & 7.20 \(\pm\) 0.77 & 7.00 \(\pm\) 0.00 & 5.27 \(\pm\) 2.25 \\
 \midrule
\multicolumn{9}{c}{\textbf{Pause-Tuned Model with Standard Pause Tokens (Technique 5)}} \\ \midrule
LLaMA 3.2 3B & 7.00 \(\pm\) 3.59 & 5.80 \(\pm\) 2.73 & 4.80 \(\pm\) 4.00 & 2.80 \(\pm\) 2.96 & 1.20 \(\pm\) 4.93 & 1.00 \(\pm\) 0.00 & 1.00 \(\pm\) 0.00 & 1.00 \(\pm\) 0.00 \\
LLaMA 3.1 8B & 8.33 \(\pm\) 2.32 & 7.93 \(\pm\) 2.66 & 8.00 \(\pm\) 2.45 & 7.07 \(\pm\) 2.46 & 6.00 \(\pm\) 2.07 & 5.80 \(\pm\) 2.34 & 4.73 \(\pm\) 2.58 & 4.33 \(\pm\) 2.66 \\ \bottomrule
\end{tabular}
}
\caption{Results of the multiple-needle task, presented as mean $\pm$ standard deviation, comparing the baseline, pause token methods, and pause-tuning. Token counts are denoted as 1K = 1,000 tokens, 2K = 2,000 tokens, etc.}
\label{tab:combined_results_2}
\end{table*}
\end{document}